\documentclass{Interspeech}
\usepackage{subfig}
\interspeechcameraready

\title{Improving Bird Classification with Primary Color Additives}

\author[affiliation={1}]{Ezhini}{ Rasendiran R}
\author[affiliation={2}]{Chandresh Kumar}{Maurya}
\affiliation{Department of Metallurgical Engineering and Materials Science}{Indian Institute of Technology Indore}{India}
\affiliation{Department of Computer Science \& Engineering}{Indian Institute of Technology Indore}{India}
\email{mems210005019@alum.iiti.ac.in, chandresh@iiti.ac.in}
\keywords{Audio Classification, Bird Classification, BirdCLEF-2024, EfficientNet}

\usepackage{comment}

\begin{document}

\maketitle

\begin{abstract}
    
We address the problem of classifying bird species using their song recordings, a challenging task due to environmental noise, overlapping vocalizations, and missing labels. Existing models struggle with low-SNR or multi-species recordings. We hypothesize that birds can be classified by visualizing their pitch pattern, speed, and repetition—collectively called \textbf{motifs}. Deep learning models applied to spectrogram images help, but similar motifs across species cause confusion. To mitigate this, we embed frequency information into spectrograms using primary color additives. This enhances species distinction, improving classification accuracy. Our experiments show that the proposed approach achieves statistically significant gains over models without colorization and surpasses the BirdCLEF 2024 winner, improving F1 by \textbf{7.3\%}, ROC-AUC by \textbf{6.2\%}, and CMAP by \textbf{6.6\%}. These results show the effectiveness of incorporating frequency information via colorization.

\end{abstract}

\section{Introduction}
Audio classification is the process of categorizing audio recordings into predefined classes based on their acoustic characteristics \cite{hershey2017cnn}. This technique is increasingly being used in biodiversity conservation efforts, particularly for the monitoring of wildlife. By using advanced machine learning models and audio processing techniques, researchers can automatically identify bird calls, insect sounds, and other wildlife vocalizations \cite{kahl2021birdnet, sheikh2024birdwhisperer}. Such methods are crucial for tracking the presence, abundance, and behavior of species in their natural habitats . Conventional biodiversity monitoring methods often involve manual observation and data collection, which are time-consuming, labor-intensive, and prone to human error \cite{gregory2004bird}. The adoption of audio classification enables more efficient and accurate monitoring, providing real-time insights into ecosystem health . For instance, automated systems can analyze large datasets from bioacoustic Passive Acoustic Monitoring (PAM) machines deployed in remote  areas, making it feasible to monitor biodiversity over extended periods. By automating the identification of species through their vocalizations, aids researchers in making informed decisions to protect ecosystems and combat threats like habitat loss and climate change  This technology application is a step toward achieving global sustainability goals by ensuring the preservation of our planet’s rich biodiversity \cite{chowfin2021using}.

Deep learning breakthroughs have revolutionized bioacoustic audio classification by enabling the extraction of deep abstract features inherent in raw data \cite{lecun2015deep}; however, overlapping species sounds, a wide array of diverse vocal motifs, and striking similarities between species' calls continue to complicate robust classification.

Unsupervised source separation, such as the MixIT approach \cite{wisdom2020unsupervised}, has been employed to disentangle overlapping bird vocalizations in complex soundscapes. This method improves classification performance by isolating individual calls and enabling cleaner extraction of acoustic features. It effectively reduces interference from various sound motifs of different classes, leading to a more precise species identification \cite{denton2022improving}. However, overseparation can, on some occasions, remove essential contextual cues, diminishing the detection probability of the most prominent species. In these instances, over-separation may also fragment longer vocalizations into isolated notes that resemble calls from other species and result in misclassification.

We therefore introduce a novel feature engineering method that embeds frequency information directly into the input  mel spectogram. This enhancement enables the model to capture frequency variations within similar vocal motifs across different species. By emphasizing these underlying differences, the model is better equipped to learn and distinguish between species calls. Our experiments revealed that this strategy significantly improves the robustness of the classification. Statistical analysis confirmed that our engineered features led to meaningful performance gains compared to model without this enhancement. Our main contributions are:
\begin{itemize}
    \item Propose a novel idea of embedding frequency information into the mel  specrogram for solving the similar motif problem in the grey-scale mel spectrogram.
    \item We demonstrate through empirical study that our proposed approach is effective in handling similar motif pattern and outperforms the BirdCLEF 2024 winner model.
\end{itemize}

% \section{Related Works}
\begin{figure}
    \centering
    \includegraphics[width=\linewidth]{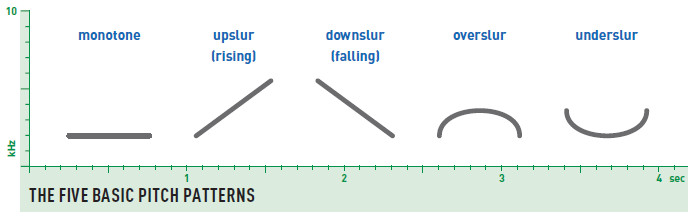}
    \caption{The five basic pitch pattern. Fig. taken from \cite{earbirding2024} with permission.}
    \label{fig:pitch-pattern}
\end{figure}
\section{Background and Problem Statement}
\subsection{Bird Sound Visualization}
Birds can be identified by visualizing their sounds \cite{earbirding2024}. Key aspects include \emph{pitch pattern, speed, repetition, pauses}, and \emph{tone quality}. Spectrogram symbols offer a simplified artistic representation of bird sounds, emphasizing patterns over detail, making them useful for humans but not for computers. For experiments, real spectrograms are used to classify bird species. Unlike music, where exact pitch matters, bird sound identification focuses on pitch variations over time. This approach enhances species recognition by capturing distinctive auditory patterns while leveraging deep learning for accurate classification. All bird sounds can be classified into five subcategories (or there combinations thereof) as shown in Fig. \ref{fig:pitch-pattern}.

\begin{figure}
    \centering
    \includegraphics[width=\linewidth, height=6cm]{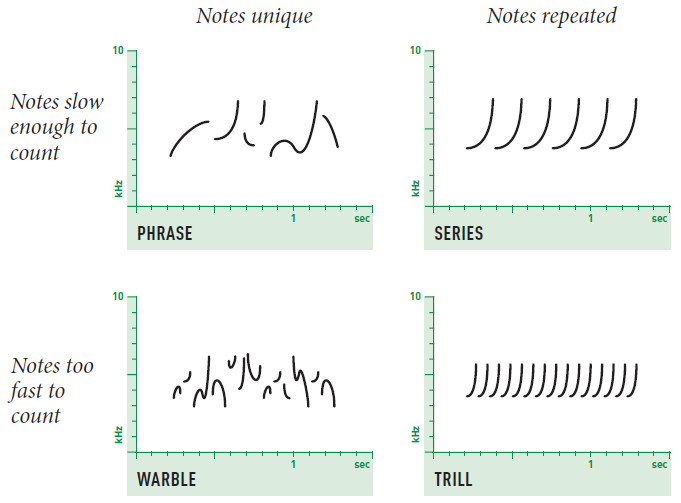}
    \caption{Repetition and speed categorization: Pharses, series, warbles, and trill. Fig. taken from \cite{earbirding2024} with permission}
    \label{fig:4-basic}
\end{figure}
\begin{figure}
    \centering
    \includegraphics[width=\linewidth]{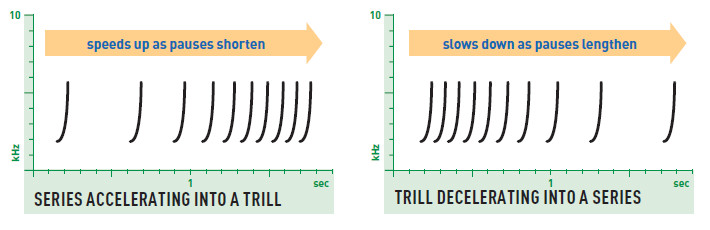}
    \caption{Series acceleration into a trill and trill de-accelerating into a series. Fig. taken from \cite{earbirding2024} with permission}
    \label{fig:acc}
\end{figure}
{\bf Monotone} sounds remain at a constant pitch and appear as horizontal lines on the spectrogram.
{\bf Upslurred} sounds increase in pitch, showing an upward tilt.
{\bf Downslurred} sounds decrease in pitch, showing a downward tilt.
{\bf Overslurred} sounds rise and then fall in pitch, with the highest point occurring in the middle.
{\bf Underslurred} sounds fall and then rise in pitch, with the lowest point occurring in the middle.

Repetition (aka motif) and speed relate to the fact that the birds sing the same note multiple times and the pace at which it happens, respectively. Together, there are four basic patterns of repetition and speed: \textbf{phraes, series, warbles} and \textbf{trills}. Phrases and series are slower sounds, where individual notes are distinct enough to count. Phrases contain unique notes that are not repeated, while series consist of a single note repeated multiple times. Warbles and trills are faster versions of phrases and series, with notes occurring too rapidly to count (typically faster than about eight notes per second).  These motifs are intricately combined to form bird songs, which vary in its pitch and pace. Temporal variations include acceleration and deceleration (shown in Fig. \ref{fig:acc}). Such dynamic patterns highlight the complexity of avian acoustics, making bird calls invaluable for species identification and behavioral studies. Bird species audio recognition ultimately comes down to identifying these unique motifs in a recording.
\subsection{Problem Statement}
Given a dataset  $\{X_i,Y_i\}_{i=1}^n$ with $X_i=[x_1,x_2, \ldots] \in \mathcal{X} $ containing multiple instances $x_j$(recordings chopped in fix-size windows as discussed in \S \ref{metho}),  and weak labels $Y_i \in \{0,1\} \in \mathcal{Y}$ at the recording labels. Here, $Y_i=1$ if any of the instance $x_i$ is positive and $Y_i=0$ if all the instances are negative. Our goal is to build a multi-class multi-label classifier  $f: \mathcal{X} \to \mathcal{Y}$.
\section{Methodology}\label{metho}
\begin{figure}
    \centering
    \subfloat[\centering Blyth's Reed Warbler]{{\includegraphics[width=0.48\columnwidth, height=4cm]{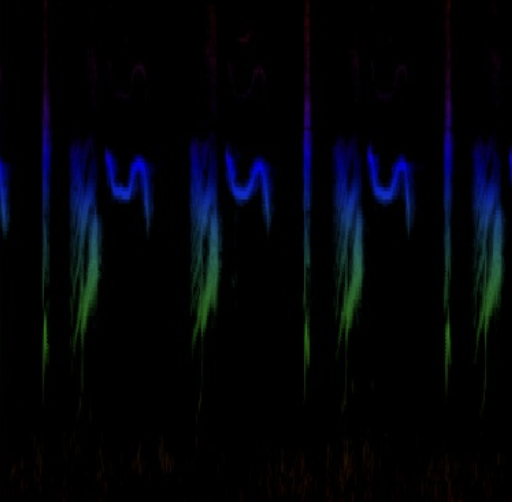} }}%
    \hfill
    \subfloat[\centering Asian Koel]{{\includegraphics[width=0.48\columnwidth, height=4cm]{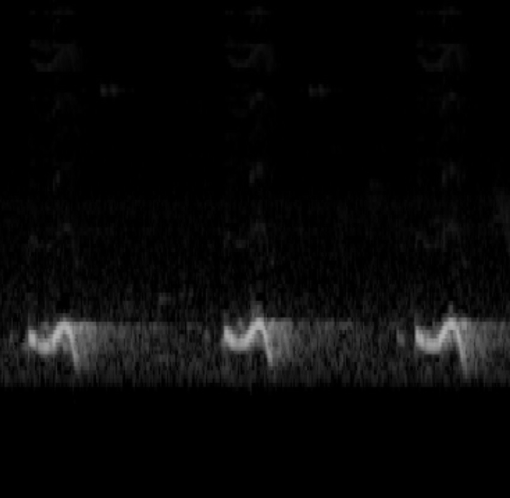} }}%
    \caption{Grayscale mel spectrogram two bird species sharing some motif patterns}%
    \label{fig:motif}%
\end{figure}

This section describes our approach to solving the audio classification of bird sounds. We hypothesize that using the mel spectrogram alone as an audio feature and feeding it to the deep learning model is insufficient. The reason is that the greyscale mel spectrogram (being a single channel) loses the frequency information (when used as an image fed to a convolution neural network (CNN)) which is a critical part of characteristic sound identification. The second issue in using the grayscale mel spectrogram to classify different bird sounds is that different birds appear to have similar motifs (series, trills, warbles, etc.) at different frequencies. For example, in Fig. \ref{fig:motif}, two different birds (Blyth’s Reed Warbler and Asian Koel) share a motif pattern. The third issue is that external sound sources like horns and birds may appear similar (in terms of motif) at the same or different frequencies.  If we somehow can embed the frequency information into the mel spectrogram, we hope to solve this problem. Next, we discuss such an approach and split our discussion into the following: (1) acoustic event detection, (2) feature engineering mel spectrogram, and (3) frequency information embedding by primary color additives. Finally, we discuss the model architecture.

% \begin{figure}
%     \centering
%     \includegraphics[height=3cm, width=\columnwidth]{images/IMG-20250114-WA0005.jpg}
%     \caption{Caption}
%     \label{fig:flameback}
% \end{figure}

\subsection{Acoustic event detection} \label{aed}
We utilize BirdCLEF 2024\footnote{https://www.kaggle.com/competitions/birdclef-2024} dataset having 182 species for classification. In this, all audio recordings are \emph{weakly labeled}, that is, a label is present at the recording level instead of the duration level. Its duration can span anywhere between 3 seconds to 30 minutes. First audio is denoised to segregate the instances of bird sound activity, and a high pass filter is applied with a cutoff frequency of 300Hz. We find that acoustic events under this threshold do not constitute any significant activity. This results in a recording that is highly populated by target acoustic events. Denoising and high pass filter reduces the energy levels of irrelevant sounds. Next, the energy is calculated for each frame as the sum of the squared absolute values of the
samples in that frame. The timings of descending energy peak value points above mean energy are calculated (using  \texttt{find\_peaks} package in \texttt{scipy.signal library}). Then a 5-second window is wrapped around the energy peaks making complete enclosure over the bird motif constituting an acoustic event. Then a multiple event level example grouping takes place with the condition that the following lower peak energy instances should not share more than 50\% time with the preceding acoustic events (to prevent overlapping acoustic events). Overall, we choose 5 events with 5 seconds duration each (sometimes it could be less number of events depending on the presence of the significant acoustic event). The probability of finding the labeled audio event among 5 sound events above mean energy is assumed as 1 in a focal
recording which means intentionally capturing sounds from a specific area of interest
compared to a PAM where the device records the entire sound without a focus. Thus, our process is guaranteed to have primary labels sound among the 5 mined sound events. 

\subsection{Feature engineering}
Next, we create a mel spectrogram of detected acoustic events. As mentioned at the beginning of this section,  mel spectrograms are single-channel and represent the pitch shift or time-stretching phenomena for different acoustic events in the recording. Computer vision models like Resnet \cite{he2016deep}, Deformable Convolutional Networks (DCNs) \cite{dai2017deformable}, etc. are translational invariance (in a sense) and hence will predict the same label for different acoustic events if motifs are shared. Alternatively, the model assigns a high probability to a secondary label as well in the case of shared motifs making the classification less accurate. Thus, the output can not be relied upon completely and we need to encode sound frequencies in some way during the learning mechanism. To this end, the mel spectrogram is normalized in the range (0,1) followed by scaling on $\log$-scale and then again normalized within the range (0,1).

\subsection{Frequency information embedding by primary color additives}
\begin{figure}
    \centering
    \subfloat[\centering Blyth's Reed Warbler]{{\includegraphics[width=0.48\columnwidth]{images/Motif-1-Frequency-encoded-Blyths-Reed-Warbler.png} }}%
    \hfill
    \subfloat[\centering Asian Koel]{{\includegraphics[width=0.48\columnwidth]{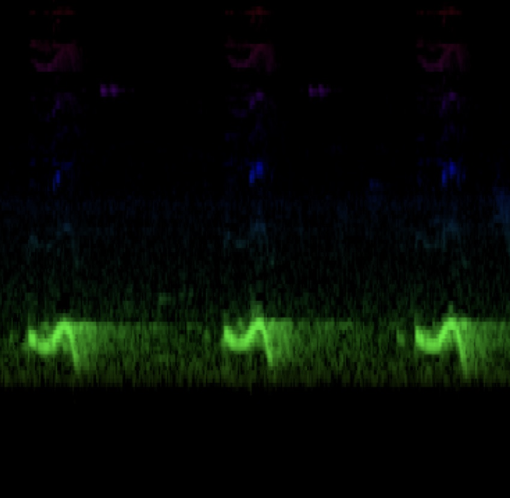} }}%
    \caption{Mel spectrogram of two bird species sharing some motif patterns. Motifs now are distinguishable by a deep learning model because of colorization.}%
    \label{fig:colormotif}%
\end{figure}

As mentioned, mel spectrogram is missing the frequency information when fed to the deep learning model. To solve this issue, we proceed as follows. The pixel representation of any mel frequency bin in an image is in an RGB channel color ratio of (1:1:1). To discriminate pixels of mel frequency bins based on their frequency information variation, the mel spectrogram is divided into 3 equal regions between
lowest frequency ($f\_min$) and highest frequency($f\_max$) initialized during mel spectrogram creation. Therefore the number of mel frequency bins in a region will be $n\_bins = total\_bins/3$ where $total\_bins$ is a parameter set during mel spectrogram creation.  The first region of the mel frequency bins whose frequency range starts from  $f\_min$ Hz its
primary color pixels will be mapped to \texttt{color\_array} RG( $1-t,t$) where $t$ is defined as
\begin{equation}
    t=\frac{\text{Index of mel frequency bin (counted from $f\_min$)}}{n\_bins}
\end{equation}
As the index of the bins increases, the red color channel linearly decreases
from pure red, while the green color channel linearly increases, both colors transitioning in equal amounts and simultaneously in every bin upwards. This primary color channel addition gives a distinct secondary color to all the mel frequency bins in a region. Finally, pixel values of the mel frequency bins are multiplied with the color array \texttt{color\_array*pixel\_value}. The same operation is performed on the mel frequency bins of other regions with the color arrays \texttt{color\_array} GB($1-t,t$) and BR($1-t,t$), respectively. As a result, we get a colorized mel spectrogram which looks like Fig. \ref{fig:colormotif}. Note that colorization may be seen as an {\it approximation} to the frequency information encoding in the mel spectrogram. As such, a colorized mel spectrogram may help distinguish two different bird species sharing a motif pattern which we show in the empirical section.

% \begin{figure}
%     \centering
%     \includegraphics[height=4cm, width=\columnwidth]{images/IMG-20250114-WA0008.jpg}
%     \caption{Colorized mel spectrograms}
%     \label{fig:colorized}
% \end{figure}

\subsection{Model Architecture} \label{modelarch}
\begin{figure}
    \centering
    \includegraphics[width=\columnwidth, height=2cm]{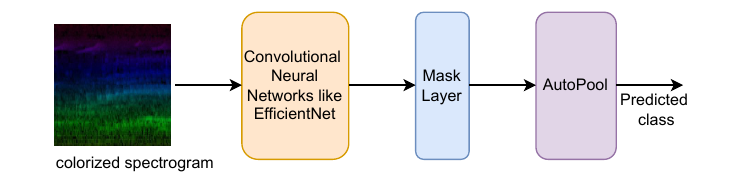}
    \caption{Model architecture}
    \label{fig:model}
\end{figure}

For audio classification, we use the EfficientNetB0 architecture which is a type of CNN \cite{tan2019efficientnet} for learning followed by the AutoPool layer \cite{10.1109/TASLP.2018.2858559} as shown in Fig. \ref{fig:model}. AutoPool is a pooling mechanism designed for weakly labeled audio classification task scenarios involving multiple-instance learning (MIL). Unlike traditional pooling methods such as max or average pooling, AutoPool introduces a trainable pooling function that learns how to aggregate instance-level predictions into recording-level predictions during training. This
flexibility allows the model to adaptively balance between max pooling (highlighting dominant
signals) and average pooling (capturing broader patterns) based on the dataset characteristics.  AutoPool is well-suited for our task because learning happens from weakly labeled data. Whereas predictions happen at a recording level during testing. AutoPool layer pools the logits of 5 instances per recording and then passes it to the sigmoid activation function for multi-class multi-label prediction. For recordings with less than five instances, zero image channels are used to fill the remaining instances, and before AutoPool, binary mask is applied on the zero image channels. 
Auto Pooled probability aggregation for multiple instances learning is given by \eqref{mil}.
\begin{equation} \label{mil}
    \hat{P}^\alpha(Y|X)=\sum_{x\in X}\hat{p}(Y|x)\left(\frac{\exp(\alpha\cdot\hat{p}(Y|x))}{\sum_{z\in X}\exp(\alpha\cdot\hat{p}(Y|z))}\right)
\end{equation}
where $\hat{P}(Y/x)$ is the probability of class $Y$ given instance $x$ and $\alpha$ is a scalar parameter learned along with the model parameters during training, which controls the pooling behavior. In particular, $\alpha=0$ equals the unweighted mean, $\alpha=1$ equals soft-max pooling, and $\alpha \to \infty$ is a max operator.
  Note that the auto pooling is done over each class separately for handling \emph{multi-label} problem. However, we do not vary $\alpha$ for each class which we leave for future study.

\subsubsection{Loss function}
The model in Fig. \ref{fig:model} is optimized for binary cross-entropy loss function \eqref{bce}.
\begin{equation} \label{bce}
\mathcal{L}_{BCE} = -\frac{1}{N}\sum_{i=1}^N \sum_{j=1}^C \left[ y_{ij}\log \hat{y}_{ij} + (1-y_{ij})\log \big(1-\hat{y}_{ij}\big) \right]
\end{equation}
where $N$ is the sample size and $C$ is the number of classes.
% \subsection{Colour mapping mel frequency bin values}

% Audio classification using computer vision models, the learning is identifying target sound patterns formed in regions of feautre image of melspectogram. The information in it is comprised of mel frequency bin values that vary across time instance. Because CV models are translational invariant they cannot learn the belonging frequency ranges  of melfrequency bin energy patterns . Look alike feautre confusion could arise when bird sounds of different species emerge in different frequency bands, but has similarity in their melfrequency bin energy pattern. Our proposed method aimed to encode the frequency range information of melfrequency bins in forms of variational secondary colors.

\section{Experiments}
\subsection{Dataset}
We utilize BirdCLEF 2024 data \cite{kahl2024overview} for training and validation of the proposed model. It has   24459 audio recording of 182 bird species of which we took only 23920 (possibly some duplicates). Due to the unavailability of the \emph{hidden} test data of BirdCLEF competition, the test set is prepared from the given audio files. We filter files with both primary and secondary labels as the secondary labels are noisy and not reliable \footnote{https://www.kaggle.com/competitions/birdclef-2024/discussion/540969}. 
% as this is a more challenging and realistic scenario in a real-world environment of passive recordings \footnote{The other kind of recording is called \emph{focal} in that the microphone is directly pointed to the bird under consideration and it usually has only the primary label.}. 
We get 1873 files which are discarded. To be consistent with the test data used in the existing literature, we use recordings with primary labels only. The remaining files are split in the ratio of 80:20 for training and validation.
% The data set taken for training and test comes from the birdclef24 training dataset, the secondary labeled test data is kept exclusively for testing, the remaining primary labeled data 80 percent of the audio files of a class is kept for training and remaining 20 percent for validation. Like wise 5 different fold of data for train and valid are prepared and trained respectively. In testing, audio is segmented into 5sec chunks and model predicts on each of them then arg max of probability from all chunks is then compared to the recording labels. The train ,valid data has 182 species and test data has 155 species.

\subsection{Baselines}
We use the winner model\footnote{https://www.kaggle.com/competitions/birdclef-2024/discussion/512197} from BirdCLEF 2024 as a baseline which is essentially an EfficientNET-B0 with some augmentations applied to the input such as horizontal cutmix, leveraging pseudo-labeled data, etc. 
Additionally, we removed the colorization from the mel spectrogram to perform an ablation study to see the effect of the color additives. 
\subsection{Training and Evaluation details}
As discussed in \S \ref{modelarch}, the model is trained following K-fold cross-validation (K=5 in our case) strategy. During validation, we feed top-5 acoustic event windows selected from the validation set following the procedure as mentioned in \S \ref{aed}. Note that this evaluation is more realistic for PAM recordings (used in BirdCLEF hidden set evaluation) compared to just testing the \textit{first} 5-second window (which may or may not contain the actual species in a PAM environment) for the presence of species (as done by the winners of the BirdCLEF).  The model is optimized with the initial learning rate of 3$e^{-3}$ decaying to 1$e^{-6}$ following cosine annealing LR scheduler. We set the batch size to 90 and epochs to 30. AdamW optimizer is used for the minimization of the loss function. The winner model is directly borrowed from the code provided by the winners \footnote{https://github.com/arpoyda/BirdCLEF\_2024}. We use the same data splits to train and test the winner model for a fair comparison. 

\subsection{Evaluation Metrics}
 Macro ROC-AUC, macro F1, and Class-averaged Mean Average Precision (CMAP) are used for the performance evaluation of the model. These metrics are best suited for multi-class, multi-label problems with imbalanced classes. In brief, CMAP is the mean of the per-class precision scores used in the previous BirdCLEF competitions \cite{kahl2019overview}. However, BirdCLEF 2024 used macro ROC-AUC. 
\subsection{Results}
As shown in Table \ref{tab:result}, we can observe that the proposed approach outperforms the winner model on all metrics by 7.3\% on F1, 6.2\% on ROC-AUC, 6.6\% on CMAP, respectively. Interestingly, we do not use any data augmentation during training/inference whereas the winner model does. To see the effect of the proposed color additives for frequency embedding, we perform an ablation study. As shown in the Table \ref{tab:result} (2nd row), we find that the colorization is effective in identifying bird specifies with similar motifs.
\begin{table}[]
\caption{Performance comparison of the proposed approach with the winner model from BirdCLEF 2024 and without colorization on the 5-fold validation set. * entry indicates a statistically significant difference from the model without colorization on the Wilcoxon signed-rank test (one-tailed) with $\alpha=0.05$. }
\label{tab:result}
\centering
\scriptsize
\begin{tabular}{|l|l|l|l|}
\hline
\textbf{Model}                                                            & \textbf{Macro-F1} & \textbf{\begin{tabular}[c]{@{}l@{}}Macro \\ ROC-AUC\end{tabular}} & \textbf{CMAP}    \\ \hline
\begin{tabular}[c]{@{}l@{}}Winner model\\ BirdCLEF\end{tabular}           & 0.6371            & 0.9220                                                            & 0.6915           \\ \hline
\begin{tabular}[c]{@{}l@{}}Proposed Model\\ w/o Colorization\end{tabular} & 0.6676            & 0.9765                                                            & 0.7217           \\ \hline
\begin{tabular}[c]{@{}l@{}}Proposed Model\\ w/ Colorization\end{tabular}  & \textbf{0.6833*}  & \textbf{0.9797*}                                                  & \textbf{0.7374*} \\ \hline
\end{tabular}
\end{table}

\section{Conclusion and Future Works}
The present work studies bird clasfication problem through the lens of multiple instance learning. To tackle the problem of identifying bird species with similar motifs pattern, we propose the idea of color additives for frequency embedding. The empirical results reveal that the colorization is an effective method for classifying birds with similar motifs at different frequency bands. There are several open research directions to proceed. We hypothesize that the colorization will be more effective for classifying overlapping vocalization combined with the voice activity detection module for audio segmentation \cite{gu2024positive}. Another interesting study may be to see the effect of the $\alpha$ parameter in the multi-class multi-label problem.
\section{Limitations}
Our study is limited to identify primary labels. As such, our model can not classify unknown bird specifies present in the audio recordings.

\bibliographystyle{IEEEtran}
\bibliography{mybib}

\end{document}